\pdfoutput=1

\documentclass[11pt]{article}

\usepackage{ACL2023}

\usepackage{times}
\usepackage{latexsym}
\usepackage{amsmath,amsfonts,amssymb}
\usepackage{graphicx}

\usepackage[T1]{fontenc}

\usepackage[utf8]{inputenc}

\usepackage{microtype}

\usepackage{inconsolata}

\usepackage[textsize=scriptsize]{todonotes}

\title{BERTScoreVisualizer: A Web Tool for Understanding Simplified Text Evaluation with BERTScore}

\author{Sebastian Jaskowski \\
  Georgia Institute of Technology \\
  \texttt{sjaskowski3@gatech.edu} \And
  Sahasra Chava \\
  Fulton Science Academy \\
  \texttt{siyasahasra@gmail.com} \And
  Agam Shah \\
  Georgia Institute of Technology \\
  \texttt{ashah482@gatech.edu}
  }

\begin{document}
\maketitle
\begin{abstract}

The BERTScore metric is commonly used to evaluate automatic text simplification systems. However, current implementations of the metric fail to provide complete visibility into all information the metric can produce. Notably, the specific token matchings can be incredibly useful in generating clause-level insight into the quality of simplified text. We address this by introducing BERTScoreVisualizer, a web application that goes beyond reporting precision, recall, and F1 score and provides a visualization of the matching between tokens. We believe that our software can help improve the analysis of text simplification systems by specifically showing where generated, simplified text deviates from reference text. We host our code and demo on GitHub:  \url{https://github.com/Sebiancoder/BERTScoreVisualizer}.

\end{abstract}

\section{Introduction}

Evaluating machine-generated text in text simplification systems remains a challenging task \cite{grabar2022evaluation}. Human evaluation remains expensive and time-consuming. As such, a number of reference-based automatic evaluation metrics have emerged, such as BLEU \cite{papineni-etal-2002-bleu}, ROUGE \cite{lin-2004-rouge}, and BERTScore \cite{zhang2020bertscoreevaluatingtextgeneration}. These metrics work by comparing generated text to some reference text. BERTScore notably improves upon the others by making use of contextual embeddings from BERT \cite{devlin2019bertpretrainingdeepbidirectional} to improve unit matching.

In addition to providing overall precision, recall, and F1 score values, the BERTScore metric can provide a matching metric (Cosine similarity between embeddings) between tokens in the reference text and the candidate text. With BERTScoreVisualizer, we aim to provide a tool that allows a user to visualize and understand these matching, allowing for greater insights into the behavior and performance of text simplification systems.

\section{Background}

\paragraph{Evaluation of Text Simplification}
Automatic evaluation approaches for text simplification models range from word-counting based approaches such as FKGL \cite{Kincaid1975DerivationON} and SMOG \cite{d9397c09-9d7e-3784-b191-6efaa0fd35d0} to more recent reference-based metrics like BLEU \cite{papineni-etal-2002-bleu}, ROUGE \cite{lin-2004-rouge}, and BERTScore \cite{zhang2020bertscoreevaluatingtextgeneration}. Among the reference-based metrics, BERTScore presents a key advantage by not relying on exact token matching, allowing it to more fairly evaluate semantically similar but syntactically different text \cite{zhang2020bertscoreevaluatingtextgeneration}.

\paragraph{BERTScore}

BERTScore evaluates a string of generated, or candidate, text against a provided reference text by utilizing the BERT language model \cite{devlin2019bertpretrainingdeepbidirectional} to generate vector representation embeddings of each token. As such, the $k$ reference text tokens are transformed into a sequence of $k$ vector embeddings  $x = \left<x_1, x_2, \dots x_k\right>$ and the $m$ candidate text tokens to a sequence of $m$ vector embeddings $\hat{x} = \left<\hat{x_1}, \hat{x_2}, \dots \hat{x_m}\right>$. The pairwise cosine similarity scores between all $km$ possible (reference token, candidate token) pairs are then calculated \cite{zhang2020bertscoreevaluatingtextgeneration}.

For every reference token, the candidate token with the highest BERT cosine similarity is identified. This cosine similarity is taken as a "recall" metric, effectively identifying how well the reference token is recalled by the best-match candidate token. An overall recall score is then calculated by averaging the individual recall scores across all reference tokens.

\begin{figure*}[t!]
    \includegraphics[width=\textwidth]{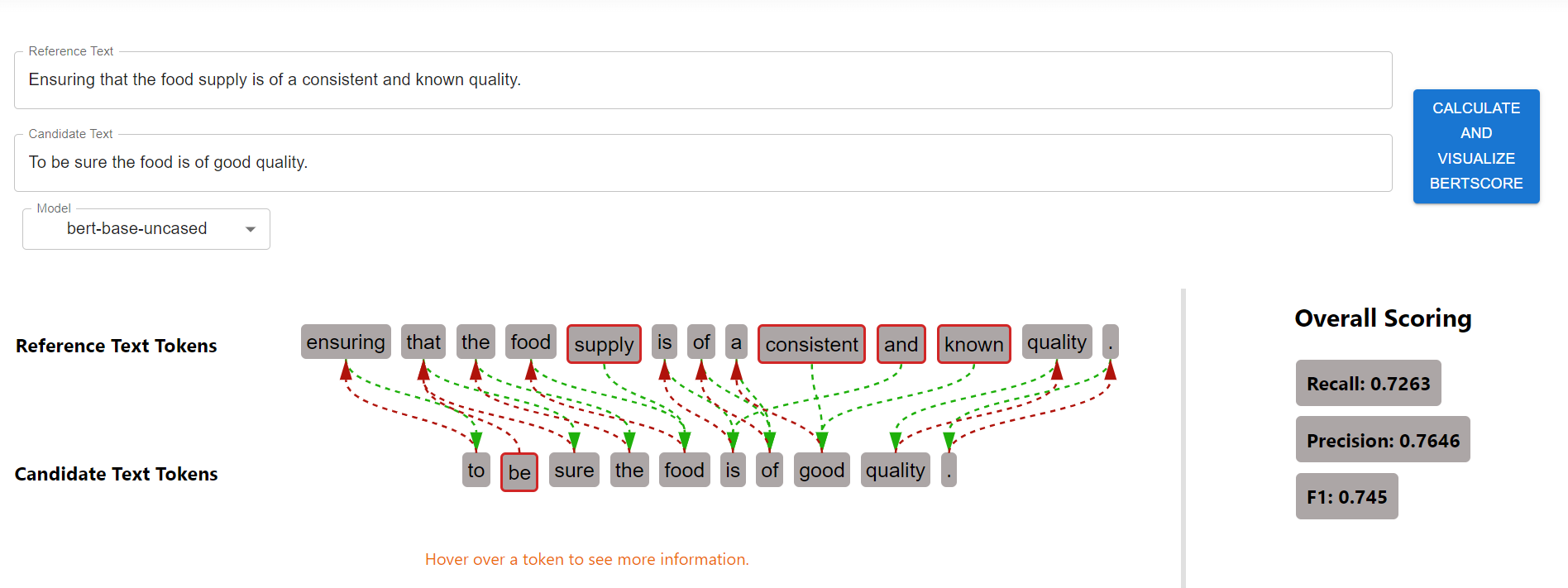}
    \caption{A screenshot of BERTScoreVisualizer, displaying the matching graph between two text sequences that have similar meanings, but are not exactly the same.}
    \label{fig:screenshot}
\end{figure*}

\begin{align}
    R_{BERT} &= \frac{1}{|x|}\sum_{x_i \in x} \max_{\hat{x_j} \in \hat{x}} \frac{x_{i}^{\intercal}\hat{x_j}}{|x_i||\hat{x_j}|}
\end{align}

A similar approach is taken to find the precision of a given candidate token, which can be thought of as a measure of how well the candidate token relates to the meaning of the reference text.

\begin{align}
    P_{BERT} &= \frac{1}{|\hat{x}|}\sum_{\hat{x_j} \in \hat{x}} \max_{x_i \in x} \frac{x_{i}^{\intercal}\hat{x_j}}{|x_i||\hat{x_j}|}
\end{align}

An overall F1 score can then be calculated as follows \cite{zhang2020bertscoreevaluatingtextgeneration}.

\begin{align}
    F_{BERT} = 2\frac{P_{BERT} * R_{BERT}}{P_{BERT} + R_{BERT}}
\end{align}

\section{System Design Goal}

BERTScoreVisualizer aims to provide an interface to visualize the matching results between candidate text tokens and reference text tokens in the BERTScore metric. As shown in Figure \ref{fig:screenshot}, for every reference token, it highlights the candidate token that best captures, or recalls, the information present in that reference token. For every candidate token, it identifies the reference token to which it best contributes in capturing semantic meaning.

This matching information can help convey insights into simplified text quality that cannot be conveyed by just the sequence-level metrics of precision, recall, and F1 score. For example, it may help the user identify that a particular sentence or clause in the reference text is not captured well by the candidate text, demonstrating that the text simplification caused a loss of information. Alternatively, the tool could help detect a particular sentence or clause in the candidate text that does not contribute to capturing the meaning of the reference text, signalling that the text was not sufficiently simplified.

\section{Notable Features}

To support effective visualization, when a token is hovered over, all other dotted, matching lines except the ones connected to that token are blurred out. This allows the user to easily see the role that the hovered token is playing in the BERTScore algorithm. In addition, a popup appears that displays the token-specific recall or precision score, with the color of the popup corresponding to the quality of the matching.

BERTScore also identifies unmatched tokens, which are tokens that are not selected as a best-match by any token in the opposing text sequence, by highlighting them in a red box. Unmatched reference tokens may signify that the simplified text output lost some of the original meaning. Unmatched candidate tokens may be interpreted as "extra" tokens that do not contribute to capturing the meaning of the reference text, possibly signaling that the text simplification system did not simplify the text sufficiently.

\section{Implementation}

BERTScoreVisualizer's frontend is implemented as a React Web Application; allowing for versatility and easy use, as it can run in a web browser. The backend is deployed as a containerized flask application, allowing for easy scalability. The backend is responsible for performing BERT inference, and running the BERTScore algorithm, while all visualization happens on the frontend. Currently, we only use the "bert-base-uncased" model for creating vector embeddings but our system can support different embedding models besides BERT. 

\bibliography{anthology,custom}

\begin{thebibliography}{7}
\expandafter\ifx\csname natexlab\endcsname\relax\def\natexlab#1{#1}\fi

\bibitem[{Devlin et~al.(2019)Devlin, Chang, Lee, and Toutanova}]{devlin2019bertpretrainingdeepbidirectional}
Jacob Devlin, Ming-Wei Chang, Kenton Lee, and Kristina Toutanova. 2019.
\newblock \href {http://arxiv.org/abs/1810.04805} {Bert: Pre-training of deep bidirectional transformers for language understanding}.

\bibitem[{Grabar and Saggion(2022)}]{grabar2022evaluation}
Natalia Grabar and Horacio Saggion. 2022.
\newblock Evaluation of automatic text simplification: Where are we now, where should we go from here.
\newblock In \emph{Traitement Automatique des Langues Naturelles}, pages 453--463. ATALA.

\bibitem[{Kincaid et~al.(1975)Kincaid, Fishburne, Rogers, and Chissom}]{Kincaid1975DerivationON}
Peter Kincaid, Robert~P. Fishburne, Richard~L. Rogers, and Brad~S. Chissom. 1975.
\newblock \href {https://api.semanticscholar.org/CorpusID:61131325} {Derivation of new readability formulas (automated readability index, fog count and flesch reading ease formula) for navy enlisted personnel}.

\bibitem[{Laughlin(1969)}]{d9397c09-9d7e-3784-b191-6efaa0fd35d0}
G.~Harry~Mc Laughlin. 1969.
\newblock \href {http://www.jstor.org/stable/40011226} {Smog grading-a new readability formula}.
\newblock \emph{Journal of Reading}, 12(8):639--646.

\bibitem[{Lin(2004)}]{lin-2004-rouge}
Chin-Yew Lin. 2004.
\newblock \href {https://aclanthology.org/W04-1013} {{ROUGE}: A package for automatic evaluation of summaries}.
\newblock In \emph{Text Summarization Branches Out}, pages 74--81, Barcelona, Spain. Association for Computational Linguistics.

\bibitem[{Papineni et~al.(2002)Papineni, Roukos, Ward, and Zhu}]{papineni-etal-2002-bleu}
Kishore Papineni, Salim Roukos, Todd Ward, and Wei-Jing Zhu. 2002.
\newblock \href {https://doi.org/10.3115/1073083.1073135} {{B}leu: a method for automatic evaluation of machine translation}.
\newblock In \emph{Proceedings of the 40th Annual Meeting of the Association for Computational Linguistics}, pages 311--318, Philadelphia, Pennsylvania, USA. Association for Computational Linguistics.

\bibitem[{Zhang et~al.(2020)Zhang, Kishore, Wu, Weinberger, and Artzi}]{zhang2020bertscoreevaluatingtextgeneration}
Tianyi Zhang, Varsha Kishore, Felix Wu, Kilian~Q. Weinberger, and Yoav Artzi. 2020.
\newblock \href {http://arxiv.org/abs/1904.09675} {Bertscore: Evaluating text generation with bert}.

\end{thebibliography}
\bibliographystyle{acl_natbib}

\end{document}